\pdfoutput=1

\documentclass[11pt]{article}

\usepackage{acl}

\usepackage{times}
\usepackage{latexsym}
\usepackage{todonotes}

\include{pythonlisting}
\usepackage{multirow, makecell}
\usepackage{microtype}
\usepackage{graphicx}
\usepackage{amsmath,amssymb,amsfonts}
\usepackage{textcomp}
\usepackage{times}
\usepackage{latexsym}
\usepackage{kotex}
\usepackage{algorithm}
\usepackage[noend]{algpseudocode}
\algrenewcommand\Return{\State \algorithmicreturn{} }
\usepackage{ctable}
\usepackage{listings}
\usepackage{enumitem}
\usepackage{xcolor}
\usepackage{placeins}
\usepackage{amssymb,amsmath}
\usepackage{cellspace}
\usepackage{tipa}

\makeatletter
\renewcommand*{\@fnsymbol}[1]{\ensuremath{\ifcase#1\or \dagger\or *\or \ddagger\or
   \mathsection\or \mathparagraph\or \|\or **\or \dagger\dagger
   \or \ddagger\ddagger \else\@ctrerr\fi}}
\makeatother

\usepackage[T1]{fontenc}

\usepackage[utf8]{inputenc}

\usepackage{microtype}

\title{\textsc{PicTalky}: Augmentative and Alternative Communication for\\ Language Developmental Disabilities}

\author{Chanjun Park$^{1,2}$\thanks{\hspace*{0.5em}{All authors contributed equally.}}, Yoonna Jang$^{1\dagger}$, Seolhwa Lee$^{3\dagger}$, Jaehyung Seo$^{1\dagger}$, Kisu Yang$^{4}$, Heuiseok Lim$^{1}$\thanks{\hspace*{0.5em}{Corresponding author.}}\\
  $^{1}$Korea University, South Korea \\
  $^{2}$Upstage, South Korea \\
  $^{3}$University of Copenhagen, Denmark \\
  $^{4}$VAIV corp, South Korea \\
  \texttt{\{bcj1210, morelychee, seojae777, limhseok\}@korea.ac.kr} \\
  \texttt{chanjun.park@upstage.ai, sele@di.ku.dk, ksyang@vaiv.kr} \\}

\begin{document}
\maketitle
\begin{abstract}
Children with language disabilities face communication difficulties in daily life. They are often deprived of the opportunity to participate in social activities due to their difficulty in understanding or using natural language. In this regard, Augmentative and Alternative Communication (AAC) can be a practical means of communication for children with language disabilities. In this study, we propose \textsc{PicTalky}, which is an AI-based AAC system that helps children with language developmental disabilities to improve their communication skills and language comprehension abilities. \textsc{PicTalky} can process both text and pictograms more accurately by connecting a series of neural-based NLP modules. Additionally, we perform quantitative and qualitative analyses on the modules of \textsc{PicTalky}. By using this service, it is expected that those suffering from language problems will be able to express their intentions or desires more easily and improve their quality of life. We have made the models freely available alongside a demonstration of the web interface \footnote{\url{http://nlplab.iptime.org:9062/}}. Furthermore, we implemented robotics AAC for the first time by applying \textsc{PicTalky} to the NAO robot. 
\end{abstract}

\section{Introduction}

The majority of people with language disabilities suffer in their daily lives as they cannot understand or speak the language. As it is a means of communication, they may be deprived of the opportunity to participate in social activities. Also, they may experience financial difficulties. In general, people with speech disorders have lower employment rates than people with other types of disabilities. What is worse is that the proportion of people with autism disorder has been increasing every year \cite{zablotsky2019prevalence}. Accordingly, a solution is required to ensure their economic freedom.

Meanwhile, augmentative and alternative communication (AAC) has been suggested and applied to solve communication problems for people with language disabilities~\cite{beukelman1998augmentative}. This approach enables nonverbal communication by replacing language. Although several AAC software resources are available, existing software packages are expensive, difficult to use, and only provide simple functions. To address these problems, we present a novel AAC system for children with language developmental disabilities. We refer to our AAC software as \textsc{PicTalky}. Neural-based grammar error correction (GEC) and a symbol-based text-to-pictogram (TP) module are utilized in our model. Thus, \textsc{PicTalky} offers neural- and symbol-based AAC for the improvement of communication and language learning, which have not been adopted in existing software. 
 
From the perspective of NLP, the speech errors from people with language disabilities can be interpreted as grammatical errors at the morphological and syntactic levels. To handle these errors, neural GEC is applied in \textsc{PicTalky}. Moreover, we consider both text and image processing for AAC education and communication. After a sentence is entered as an input through the speech-to-text (STT) module, it is passed through the neural GEC and natural language understanding (NLU) modules. Finally, the corresponding pictograms are displayed. 

\textsc{PicTalky} is aimed at children aged 0 to 14 years who have language developmental disabilities caused by intellectual or autism disabilities. The first reason that we focus on children is that early treatment during childhood is critical. According to~\citet{lenneberg1967biological}, language must be acquired during a critical period that ends at approximately the age of puberty with the establishment of the cerebral lateralization of function. Unless language is learned during this period, it is difficult for language to be used freely. This may result in social deterioration, contraction, aggression, and other problematic behaviors, which eventually affect the overall quality of life and satisfaction of the person~\cite{schwarz2001diagnosis}.

The second reason is that there is currently insufficient social support for language therapy. Not all children with developmental disabilities can benefit from public systems owing to the limited support. Moreover, in addition to the children, their family and caregivers them experience difficulties. 

Therefore, we propose \textsc{PicTalky}, which complements the limitation of existing products and increases the accessibility of children with language disabilities to appropriate education and treatment. We expect that not only the people with language disabilities but also their caregivers can have more easier education and communication by using this service. Furthermore, in addition to the implementation of the web application, we apply \textsc{PicTalky} to the NAO robot, thereby providing the first robotics AAC. We expect that robotics AAC can draw interest of children, so that they can use AAC more friendly and easily.

Our contributions are as follows:

\begin{itemize}
\item { We propose \textsc{PicTalky} for people with language diabilities, which is the first AAC software with GEC and a synonym-replacement system for accurate language processing. }

\item{ We analyze each detailed function of \textsc{PicTalky} quantitatively and qualitatively. Also, we measure the satisfaction score during the actual services. }

\item{ We present a novel metric known as text-to-pictogram accuracy (TPA) to measure the performance of converting texts into pictograms. }

\item { We open \textsc{PicTalky} in the form of a platform, so that it can help people with language disabilities and contribute to the research in this area. }

\item{We implement robotics AAC for the first time by applying \textsc{PicTalky} to the NAO robot.}
\end{itemize} 

\begin{figure*}  
  \centering
  \includegraphics[scale=0.50]{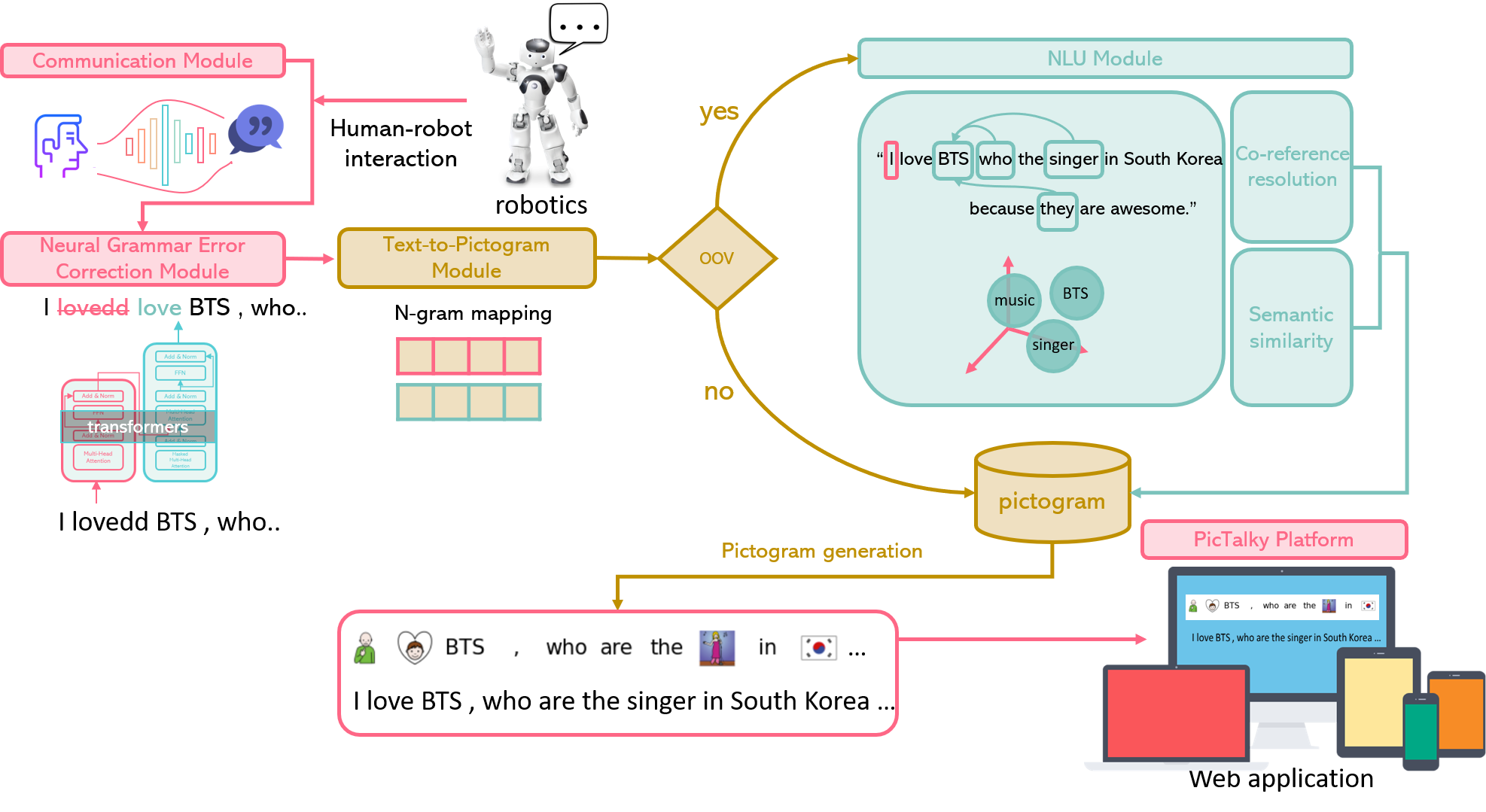}
  \caption{Overall architecture of \textsc{PicTalky}. }
  \label{fig:boat1}
\end{figure*}

\section{Background}
\subsection{AAC Software for Language Developmental Disabilities}
Several AAC software platforms have been developed for language education. TouchChat\footnote{\url{https://touchchatapp.com/}} is a symbol- and text-based AAC tool with a text-to-speech (TTS) service. AVAZ\footnote{\url{https://www.avazapp.com/}} is a language education service that uses pictograms. TalkingBoogie~\cite{shin2020talkingboogie} is software that supports the caregivers of children.

Systems that use AAC have also been developed for communication in daily life. Proloquo2Go\footnote{\url{https://www.assistiveware.com/products/proloquo2go/}} and QuickTalkAAC\footnote{\url{https://digitalscribbler.com/quick-talk-aac/}} enable people to communicate by using symbols or text with a TTS service. iCommunicate\footnote{{\url{http://www.grembe.com/}}} is a visual and text AAC application that allows for the creation of pictures and storyboards. Although several AAC software platforms have been developed, certain problems remain, such as difficulty of use and high costs. Moreover, existing pictogram-based AAC software is difficult for users to use because they need to select an image from the communication board by themselves. 

\textsc{PicTalky} is the first symbol-based AAC system with neural GEC to provide more accurate and sophisticated language education and communication. \textsc{PicTalky} automatically outputs the sequence of the pictograms according to the spoken sentences. It can be used for communication between people with disabilities as well as between people with disabilities and non-disabled people. Moreover, it offers the potential to be extended to multilingual versions by using neural machine translation.

\subsection{Symbolic AAC}
AAC enables nonverbal communication instead of a language, and it can provide practical help for people with cognitive and linguistic disorders.

In the majority of studies on AAC, researchers have employed graphic symbols (i.e., pictograms and picture communication symbols)~\cite{kang2019cultural} as alternative means of language items to improve the communication skills of children with language developmental disabilities. In this manner, children can be taught how to express their needs and interact with others using symbols~\cite{huang2019effects}.

Most authors have claimed that graphic symbols can enhance the literacy skills and communication of children or support children with disabilities in functional competence (e.g., writing, improving their communication partner knowledge, and learning)~\cite{karal2016standardization,nam2018overview, light2019challenges}.
Finally, AAC software is a form of symbolic knowledge representation~\cite{beuke2013}. That is, symbols are verbal or visual representations of ideas and concepts. Therefore, we adopt both text and image processing mechanisms (i.e., TP) to consider symbolic knowledge with NLP in AAC. 
Furthermore, we use a deep learning architecture approach for our GEC module. To the best of our knowledge, no such method for a neural and symbol mechanism in AAC has yet been presented.

\section{\textsc{PicTalky}}
\subsection{Communication Module}
Our proposed service uses deep learning-based speech-to-text (STT), which takes the voice of the user as input and converts it into text. We adopt Naver CLOVA Speech \cite{chung2019naver} for the STT system. The text input can be entered with the keyboard as well as in the form of voice. Users and caregivers can enter the text input easily with the keyboards of their personal computer, tablet, or mobile phone. 

\subsection{Neural GEC Module}
People with language disabilities tend to make grammar and pronunciation errors when speaking. The grammar error correction (GEC) system revises various linguistic errors of users, so it is useful for children to practice correct sentences.

PicTalky is equipped with a neural GEC module that accurately corrects the STT outputs. We denote the sequence-to-sequence model that is applied to the GEC task as neural GEC. From the perspective of machine translation, the neural GEC task is a system whereby a sentence with noise and a correct sentence are entered as the source and target sentences, respectively. Subsequently, translation from the input to the output is trained with the sequence-to-sequence model. In this method, training is conducted without specifying a particular error type; thus, various errors can be detected and processed simultaneously. 

\textsc{PicTalky} enhances the software quality with the latest GEC technique which utilizes noising encoder and denoising decoder proposed by~\citet{park2020neural} with copy mechanism ~\cite{gu2016incorporating}. As a result, the speech errors of people with developmental disorders can be corrected on the text level. 

\subsection{TP Module}
Pictograms are complementary and alternative means of communication that can help people with language difficulties. Unlike languages, which require an understanding of rules and symbolic systems, pictograms deliver the meaning more intuitively and rapidly. Thus, pictograms are utilized in the language rehabilitation field. For example, by using pictograms on communication boards, children can learn how to communicate with others~\cite{calculator1983evaluating}. 
Pictograms provide children who have not learned the language system with practical help in language comprehension and speaking.

This study presents a system that causes the output of the pictogram images to correspond to the input text by using text and image processing. The text-to-pictogram (TP) module is an N-gram base mapping system, and it returns the output images that are morphologically similar to the input text in the pictogram dataset. The pictogram dataset includes texts such as words, phrases or sentences that explain the corresponding images. For more accurate mapping, our TP module makes use of a method that scans the entire sentence by N-gram to 1-gram and provides the most similar image. 

\subsection{NLU module}
The output of the TP module is processed by the natural language understanding (NLU) module to handle the out-of-vocabulary (OOV) text that is not in the pictogram dataset. For this reason, we propose a method that causes the input vocabulary to correspond to a semantically similar image. 

In the NLU module, unknown words are replaced with substitute words by measuring the semantic similarities, and a co-reference resolution system is applied to the substitute words. The semantic similarities are measured by Word2Vec~\cite{mikolov2013efficient} and WordNet~\cite{miller1995wordnet}. Within the input text, substitute words can be resolved through the co-reference resolution function of the spaCy\footnote{\url{https://spacy.io/}} library. The remaining grammatical elements, such as unknown vocabularies, conjunctions, and articles that are not processed by measuring the semantic similarity and replacing unknown words with substitute words are designed not to be printed in the output image.

\subsection{Overall Architecture of \textsc{PicTalky}}\label{sec:overall}
When voice input is entered, it is converted into text by the communication module. Subsequently, the text is corrected by the neural GEC system and the corrected texts are changed into pictograms using the TP module. If OOV text exists in the input, the NLU module addresses this problem. Finally, a corresponding pictogram sequence is output. 

The overall structure of our proposed service is depicted in Figure \ref{fig:boat1}.
If an erroneous sentence \texttt{"I lovedd BTS"} is entered as input, the neural GEC corrects the input to \texttt{"I love BTS."} Eventually, the text from the pictogram is generated and this module is provided to a form of web service or robotics.

\textsc{PicTalky} aims to help children with developmental disabilities to communicate and improve their language understanding. The simultaneous encoding and transmission of speech text, both audibly and visually, allows users to understand the speaker's intentions intuitively, in spite of their difficulties in using language. Furthermore, as the text and images are delivered together, implicit learning is possible without directly teaching each element of the language. \textsc{PicTalky} is intended for children with developmental disabilities, but it can also be applied to rehabilitation for educationally disadvantaged groups.

\section{Experiment and Results}
\subsection{Datasets}
To substantiate the performance of \textsc{PicTalky} qualitatively, we adopted a test set that was provided by a GEC service company\footnote{\url{https://www.llsollu.com/}}. The test set was constructed while performing the actual GEC service, inspired by cases in which people with language developmental disabilities utter grammatically incorrect sentences. Thus, it can be stated that it provides high objectivity and reliability. We refer to this test set as the in-house test set. The test set consists of 100 sentences.

We used parallel corpora as the training data for training our neural GEC model, which were provided by Lang8~\cite{cho2013lang}. We utilized an open-source pictogram dataset that was released by the Aragonese Centre for Augmentative \& Alternative Communication\footnote{\url{http://arasaac.org}}.

\subsection{Verification of Neural GEC Module}
\paragraph{Model} Although the majority of recent NLP studies have been conducted based on the pretrain-finetuning approach (PFA), it is difficult to service a PFA-based NLP application owing to its slow speed and high computational cost, among other factors~\cite{park2021should}. Although state-of-the-art neural models such as mBART~\cite{liu2020multilingual} have been developed, the parameters and model sizes are too large to service in the industry. To overcome this problem, we built a model based on the vanilla transformer, which is easy to service. The hyperparameters were set to the same values as the settings in~\citet{vaswani2017attention}. The vocabulary size was 32,000 and sentencepiece~\cite{kudo2018sentencepiece} was adopted for the subword tokenization.

\paragraph{Performance of Neural GEC}
We used GLEU~\cite{napoles2015ground} and BiLingual Evaluation Understudy (BLEU)~\cite{papineni2002bleu} as evaluation metrics to verify the performance of the neural GEC module. GLEU is similar to BLEU, but it is a more specialized metric for the error correction system, as it considers the source sentences. The overall comparison results are presented in Table \ref{tab:my-table1}.

\begin{table}[h]
\centering
\begin{tabular}{lll}
\hline \textbf{Test set} & \textbf{BLEU}  & \textbf{GLEU} \\ \hline
In-house (\citet{park2020comparison}) &	63.77 & 53.99 \\
\hline
\end{tabular}
\caption{\label{tab:my-table1} Performance of neural GEC module.}
\end{table}

\begin{table}[h]
\centering
\scalebox{0.8}{
\begin{tabular}{ccccc}
\hline {} & \textbf{Case} & \multicolumn{2}{c}{\textbf{Deletion setting}} &
\multicolumn{1}{c}{\textbf{Score}} \\ 
{} & {} & {POS} & {Stopwords} & {} \\ \hline
 \multirow{4}{*}{\textsc{TPA}} & {(1)} &	{\checkmark} & {\checkmark} & \textbf{94.16} \\ 

{} & {(2)} &	{-} & {\checkmark} &	63.96 \\
{} & {(3)} &	{\checkmark} & {-} &	52.77 \\
{} & {(4)} &	{-} & {-} &	43.59 \\ \hline
\multirow{4}{*}{\textsc{TPA w/ penalty}} & {(1)} & {\checkmark} & {\checkmark} &	\textbf{91.62} \\

& {(2)}  &	{-} & {\checkmark} &	62.24 \\
& {(3)}  &	{\checkmark} & {-} &	51.35 \\
& {(4)}  &	{-} & {-} &	42.42 \\

\hline
\end{tabular}}
\caption{\label{tab:my-table2} Experimental results of \textsc{PicTalky}. POS represents the removal of determiners, prepositions, and conjunctions using POS tagging information.}
\end{table}

\begin{algorithm}
\small 
\caption{TPA} \label{alg:tpa_pseudocode}
\begin{algorithmic}[1]
\State Initialize $S_{pos}$ = \{determiner, preposition, conjunction\} 
\\ \scalebox{0.8}{\texttt{/* The set of exceptional POS tags */}}
\State Initialize $S_{stop}$ as stopwords predefined by NLTK
\Procedure{TPA}{$sentence$}
    \State Initialize $score$ and $N$ as zeros
    \State $W \gets PosTagger(sentence)$ 
     \\ \scalebox{0.8}{\texttt{/* Split words with POS tags */}}
    \For{each word $w \in W $}
        \If{$w.pos \notin S_{pos}$ and $w \notin S_{stop}$}
            \State $score \gets score + \delta_{\hat{y},y}$ where $\hat{y}=M_{\theta}(w)$
            \\ \scalebox{0.8}{\texttt{/* Kronecker delta of TP prediction */}}
            \State $score \gets score - (1 - \delta_{\hat{z},z})$
            where $\hat{z}=N_{\phi}(w)$
            \\  \scalebox{0.8}{\texttt{/* Penalty for a misclassified named entity */}}
            \State $N \gets N + 1$
        \EndIf
    \EndFor
    \Return $score / (N + \epsilon)$
\EndProcedure
\end{algorithmic}
\end{algorithm}

In the experimental results, BLEU and GLEU scored 63.77 and 53.99, respectively. These results are sufficiently competitive with the results of other neural GEC studies~\citep{im2017denoising,choe2019neural,park2020neural,park2020comparison}. This implies that our neural GEC module have an ability to correct the errors from the STT module, as well as the speech errors of users.

\subsection{Verification of TP Module}
The results of the performance evaluation of the TP module, which is a core function of \textsc{PicTalky}, are presented in this section. 

\paragraph{TPA} We propose text-to-pictogram accuracy (TPA), which is a novel metric for measuring the performance of the TP module. TPA is an objective indicator of how effectively the text in \textsc{PicTalky} input is converted into pictograms. The measurements are performed as follows. First, the input sentences are separated into words and POS tagged. Thereafter, the words that are POS tagged as determiners, prepositions, conjunctions (POS), and stopwords are removed, as we believe that these words are meaningless to be converted into pictograms. Thus, the words that do not contain important contents are removed during this process. The remaining words are used for the measurements and the ratio of the words that are effectively converted into pictograms is used as the TPA value. A named entity recognition (NER) penalty is also implemented when calculating the TPA value. The NER penalty is assigned when the named entities are misclassified by the NER process for the input sentences. As the named entities are important information that should be converted without errors, the NER penalty is assigned in those cases. The pseudo-code for the TPA is described in Algorithm \ref{alg:tpa_pseudocode}. 

\paragraph{Case Study} We perform comparative experiments on the TPA with various cases of deletion, as indicated in Table \ref{tab:my-table2}. There were four cases in total for the deletion cases: (1) both POS (words tagged as determiners, prepositions and conjunctions) and stopwords are deleted, (2) only stopwords are deleted, (3) only POS are deleted, and (4) neither POS nor stopwords are deleted. We also measured how the penalty affected the overall performance. We used NLTK~\cite{bird2006nltk} to remove the determiners, prepositions, conjunctions, and stopwords and used the BERT-based~\cite{devlin2018bert} NER model provided by Huggingface~\cite{wolf2019huggingface} for the penalty. 

The experimental results demonstrated that case (1) of the TPA, which was our proposed method, achieved the highest score of 94.16. In case (2) of the TPA, the score decreased by 30.20 points. When words that were POS tagged as determiners, prepositions, and conjunctions were deleted in case (3), lower performance was exhibited than in case (2). Finally, case (4) achieved the lowest performance. These results demonstrate that excluding both the POS and Stopwords from the subjects of the measurements is the most reasonable evaluation for TP conversion. Moreover, when the NER penalty was applied, the performances decreased in all cases, which means that the NER penalty contributes to more valid measurement. We also conducted a qualitative analysis on the results of \textsc{PicTalky} (see Appendix~\ref{sec:appe1}). Finally, we verify the the practicality of \textsc{PicTalky} with a questionnaire-based satisfaction survey (see Appendix~\ref{sec:appe4}.

\section{\textsc{PicTalky} with Robotics} 
We have distributed \textsc{PicTalky} as a web application (see Appendix~\ref{sec:appe2}). However, in the case of the web application, there is a possibility that it is difficult or boring for children to handle. Therefore, in addition to the web service, we have applied robotics technology to \textsc{PicTalky} for arousing interest in children. The NAO robot~\cite{shamsuddin2011humanoid,jokinen2014multimodal} is mounted in the communication module of \textsc{PicTalky}. 

NAO is the humanoid robot developed by SoftBank Robotics\footnote{\url{https://www.softbankrobotics.com/emea/en/}}. Nao has eight full-color RGB LEDs, an inertial sensor, two cameras, and many other sensors. It also has a sonar sensor to check the distance of objects in its vicinity to comprehend its environment with precision and stability. It enables NAO to react its body to move when exposed by interaction. NAO is also available in social robotics~\cite{fong2003survey}, which focuses on communicating robots capable of interacting and cooperating with humans. All of these characteristics in NAO suit our research pursuit in terms of interacting with a human.

We have created a human-robot interaction system whereby the NAO robot has a conversation with the end users and the pictograms are printed onto the connected screen. As children show substantial interest in robots, this will aid in more familiar education as opposed to web or other applications~\citep{sennott2019aac}. 
The video of our demo is also attached with our paper\footnote{\url{https://bit.ly/2SunbaW}}. 

To the best of our knowledge, this study is the first to apply \textsc{PicTalky} to the NAO robot and to develop robotics AAC for the first time. 

\section{Conclusion and Future Work}
We have proposed \textsc{PicTalky}, which is the first AI-based AAC service. With the series of deep learning-based modules, it is able to take a speech or text input from the user, correct error, and converts it into a pictogram automatically for more convenient communication and education of the people with language disabilities. The aim of our proposed system is to provide an opportunity of communication and connection among all people, without anyone being excluded. In the future, we will expand the \textsc{PicTalky} data to multilingual data for use in various languages and to make it publicly available. In addition, we plan to conduct various AI for accessibility studies to improve the quality of life for the people with disabilities. Starting with our research, we look forward to advances in many other studies so that all members of society can get benefits from AI technology without a financial burden.

\section*{Acknowledgments}
This work was supported by Institute of Information \& communications Technology Planning \& Evaluation(IITP) grant funded by the Korea government(MSIT) (No. 2020-0-00368, A Neural-Symbolic Model for Knowledge Acquisition and Inference Techniques) and by the MSIT(Ministry of Science and ICT), Korea, under the ITRC(Information Technology Research Center) support program(IITP-2022-2018-0-01405) supervised by the IITP(Institute for Information \& Communications Technology Planning \& Evaluation). In addition, This research was supported by Basic Science Research Program through the National Research Foundation of Korea(NRF) funded by the Ministry of Education(NRF-2021R1A6A1A03045425). This work is the extended version of our papers [\cite{park2020ai}]. Finally, the authors thank Yanghee Kim (rumbinie4@gmail.com) for provide initial ideas and proofreading. 

\bibliography{anthology,custom}
\bibliographystyle{acl_natbib}

\clearpage
\appendix
\onecolumn
\section{Language Developmental Disabilities}
Language disorder is a slow-speech phenomenon due to late development of the speech center in the brain \cite{tomblin2003stability}. Language disorders can be categorized into four main categories: expressive language disorder, mixed receptive-expressive language disorder, phonological disorder, and stuttering.

People with expressive language disorder have relatively normal receptive language ability to understand other people's words, but difficulty in language expression. They tend to replace simple words or sentences with gestures. People with mixed receptive-expressive language disorder shows a disability in understanding other people's words and in expressing their thoughts in language. In phonological disorder, there is a common occurrence of incorrect pronunciation in consonants, especially mispronouncing consonants or omitting the coda (auslaut) of syllables. Most frequently mispronounced consonants are \textipa{[s]}, \textipa{[z]}, \textipa{[S]}, \textipa{[Z]}, etc, also, there are mispronunciations in vowels too. The speech of people with stuttering is cut off abnormally often or the speed of which is irregular. The repetition of sounds or syllables, extension of speech sounds, and blockage of speech can be observed. Also, their speech typically begins by repeating the first consonant of a phrase. Children generally do not recognize stuttering, as getting older, they become aware of their speaking problems, and emotional reactions occur to avoid being not fluent. 

These disorders can be interpreted as grammatical errors at morphological and syntax levels from the perspective of natural language processing. Thus, deep learning-based grammar error corrector has been developed and loaded into \textsc{PicTalky}'s software .

\section{Qualitative Analysis} 
\label{sec:appe1}
We also perform a qualitative analysis on the results of \textsc{PicTalky} based on the developmental stages of the First Language Acquisition (FLA)~\cite{lightbown2021languages}. 

In Table~\ref{tab:my-table4}, the input sentences contain grammatical problems, including fronting, infinitive, article, spelling, plural -s, and irregular past form errors. The \textsc{PicTalky} web application shows users the most appropriate pictograms and the output sentences with the errors corrected. 

Examples such as \texttt{"I love play the baseball"}, \texttt{"I love danceing with a friends"}, and \texttt{"He taked my toy!"} occur in telegraphic speech \cite{chomsky1964development} in the immature language development stage between the ages of two and three. Grammatical errors for morphemes are not merely an imperfect imitation of adults' speech, and consistency of correction with frequent interactions is required to expand cognitive development. 

\texttt{"Is the dog is tired?"} and \texttt{"Do I can eat a pizza?"} are one of the errors encountered in acquiring basic structures of the first language between the age of 4 and the school years, and this stage requires correction of low frequency and complex systems. Therefore, we reproduce humans' universal language acquisition process, including frequent errors in the early and later development stages.

Note that if the Neural GEC module cannot correct grammatical errors, the NLU module can compensate for it. However, these aspects need to be supplemented through future research.

\begin{table*}[h]
\centering\resizebox{\textwidth}{!}{
\begin{tabular}{c c c}
\toprule \textbf{Input sentence} & \textbf{Output sentence} & \textbf{Pictogram}  \\ \midrule \midrule
\text{*} Is the dog is tired?
 & Is the dog tired? & \raisebox{-0.4\totalheight}{\includegraphics[width=0.35\textwidth, height=10mm]{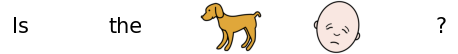}} \\

\text{*} Do I can eat a pizza? 
 & Can I eat a pizza? & \raisebox{-0.4\totalheight}{\includegraphics[width=0.38\textwidth, height=10mm]{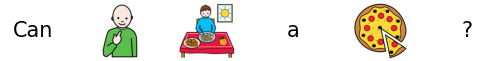}} \\

\text{*} I love play the baseball
 & I love to play baseball & \raisebox{-0.4\totalheight}{\includegraphics[width=0.4\textwidth, height=10mm]{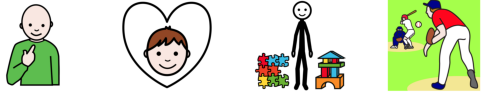}} \\

\text{*} I love danceing with a friends
 & I love dancing with friends & \raisebox{-0.4\totalheight}{\includegraphics[width=0.4\textwidth, height=10mm]{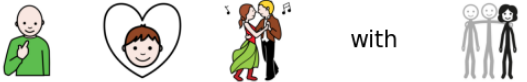}} \\

\text{*} He taked my toy!
 &	He took my toy! & \raisebox{-0.4\totalheight}{\includegraphics[width=0.4\textwidth, height=10mm]{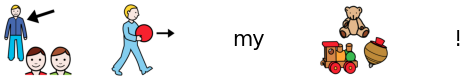}}
 \\ 
\bottomrule
\end{tabular}
} 
\caption{\label{tab:my-table4} Example sentences and pictograms for qualitative analysis created by \textsc{PicTalky} web demo.}
\end{table*}

\section{\textsc{PicTalky} Satisfaction Survey}
\label{sec:appe4}
We conducted a satisfaction survey to investigate the user satisfaction. It is difficult to employ the nonverbal child, so we identified the extreme difficulty in performing a large-scale survey. Therefore, we conducted a system satisfaction survey to 53 people with 43 experts in language disabilities and ten nonverbal children. The experts consist of thirty teachers of nonverbal children and the thirteen professionals who majored in language disabilities from Korea University Anam Hospital. \textsc{PicTalky} Satisfaction Questionnaires are shown in Table~\ref{tab:my-table3}.

\begin{figure}[tbh]  
  \centering
  \includegraphics[scale=0.5]{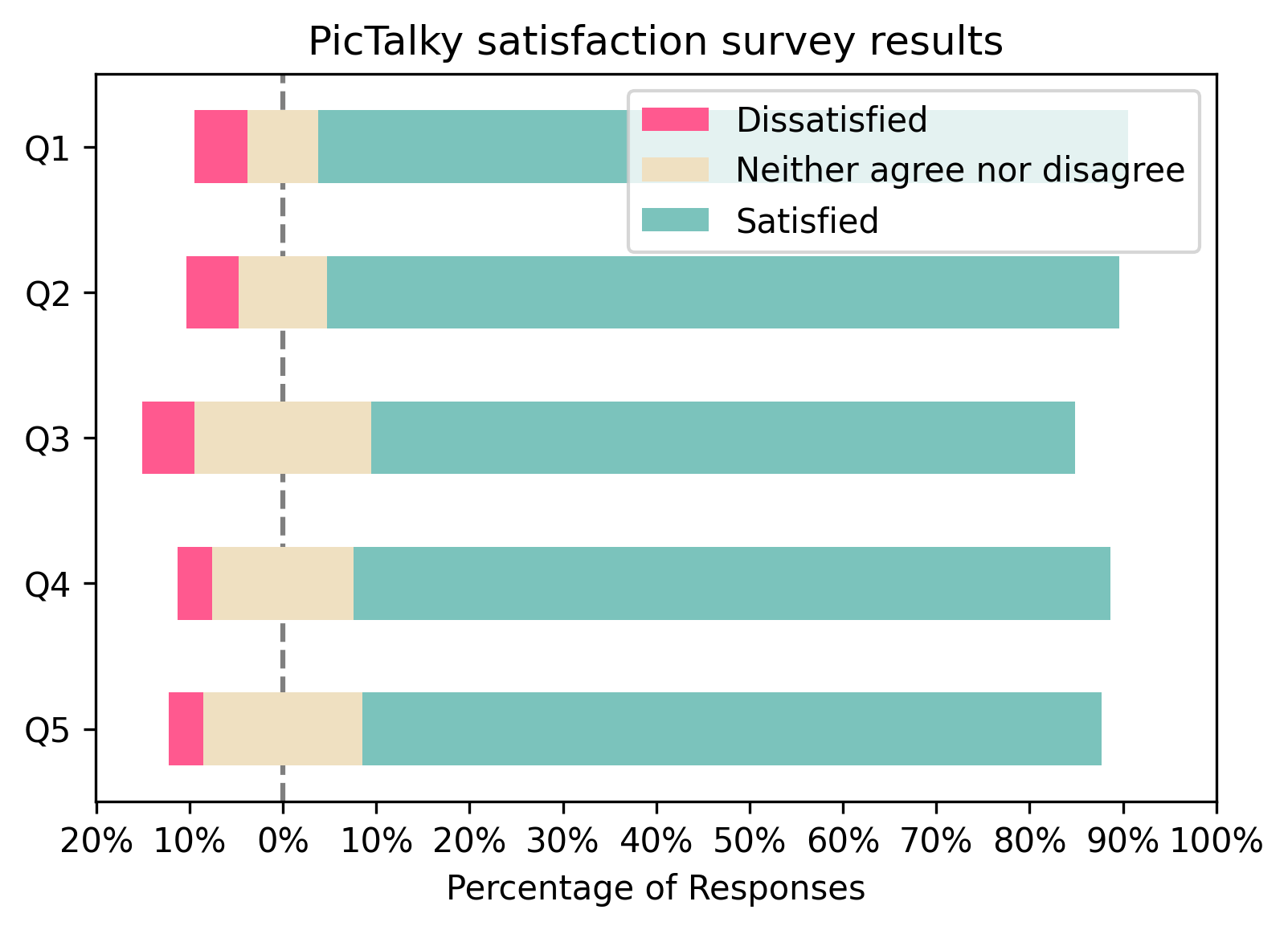}
  \caption{Response results of satisfaction survey regarding \textsc{PicTalky}. }
  \label{fig:satis}
\end{figure}

We established a total of five questions and specified the answers using a Likert scale~\cite{likert1932technique} of ``Satisfied,'' ``Neither agree nor disagree,'' and ``Dissatisfied.''. The survey results are depicted in Figure \ref{fig:satis}.

\begin{table*}[tbh]
\centering
\scalebox{0.85}{
\begin{tabular}{llllll}
\hline \textbf{Question}  \\ \hline
Q1. Are you satisfied with the overall performance of \textsc{PicTalky}? \\
Q2. Do you think this system will be helpful to people with language developmental disabilities? \\
Q3. Are you satisfied with the usability and UI of \textsc{PicTalky}? \\
Q4. Are you satisfied with the performance of the grammar error correction system? \\
Q5. Are you satisfied with the results of the text-to-pictogram function? \\
\hline
\end{tabular}}
\caption{\label{tab:my-table3} Questions of \textsc{PicTalky} satisfaction survey.}
\end{table*}

\begin{figure}  
  \centering
  \includegraphics[scale=0.48]{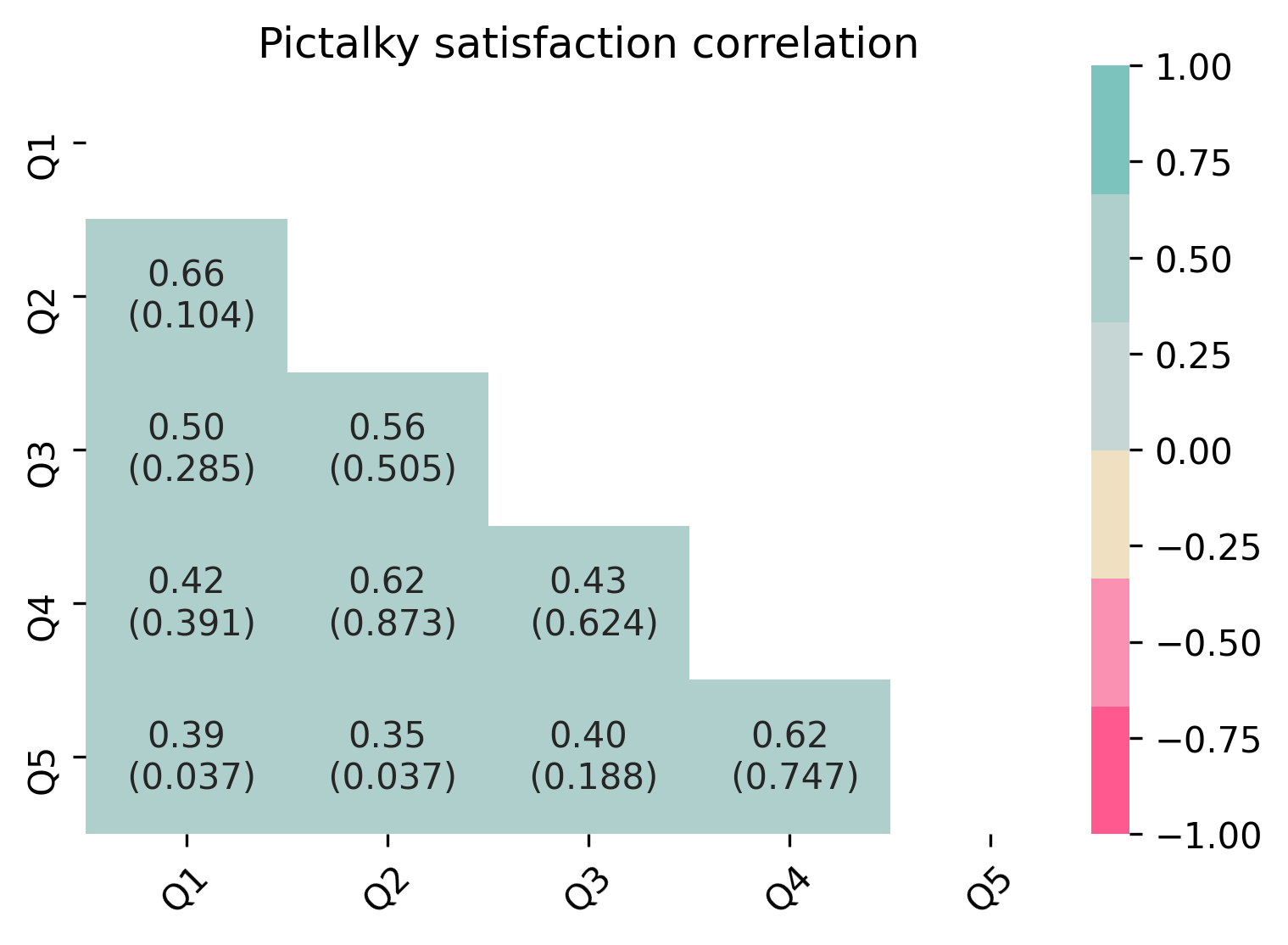}
  \caption{Results of statistical significance test using Spearman correlation between questionnaires. The weight indicates the correlation value and the value in parentheses is the p-value (p-value$<$0.05 indicates statistical significance).}
  \label{fig:satis_statistic}
\end{figure}

The survey results revealed that most people were satisfied with the performance of \textsc{PicTalky}. For each question, 80\% to 90\% of the responses were satisfied and approximately 90\% of the responses stated that it will be helpful to people with developmental disabilities. However, the UI of \textsc{PicTalky} still requires improvement and the performance of the GEC system should be enhanced. In particular, according to the results of the Spearman correlation \cite{de2016comparing} of the sentences, as illustrated in Figure \ref{fig:satis_statistic}, the correlation between Q1 and Q2 was high, which indicates that the purpose of this study was well reflected. Although the correlation between Q1 and Q5, and that between Q2 and Q5 were lower than the others, their p-values were lower than 0.05 which means the results were statistically significant.

\section{\textsc{PicTalky} with Web Application}
\label{sec:appe2}
\begin{figure}[tbh]  
  \centering
  \includegraphics[width=1.0\linewidth]{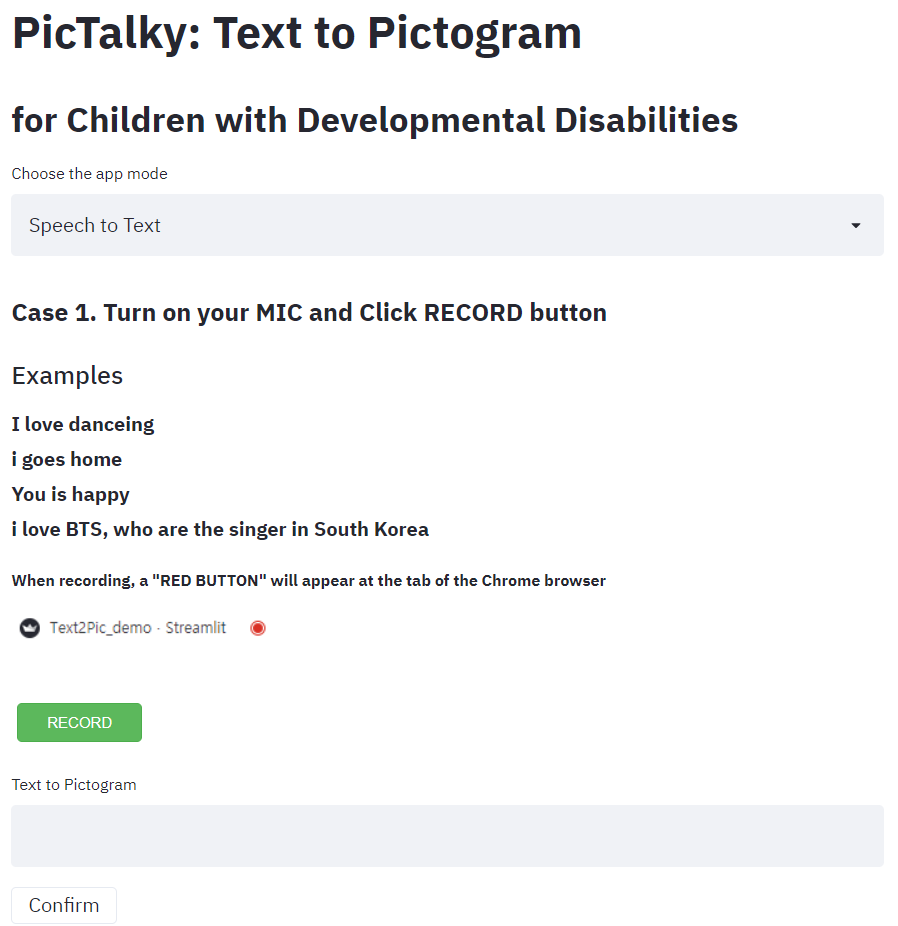}
  \caption{\textsc{PicTalky} web application.}
  \label{fig:web application}
\end{figure}

We released the \textsc{PicTalky} as the form of a web application as shown in Figure~\ref{fig:web application}. Thus, any devices enable our system by responsive user interface. The neural GEC module was connected by Rest API and distributed as both CPU and GPU services. Our system is operated by Flask under a cloud server. 

Also, we provide the user input into two modes of both speech and text considering the environment where speech is not possible. For reviewing other situations that people with language disabilities face, these settings are available to people who have deaf-mutism, aphasia. Our system is built on the compact user interface and freely available to advance accessibility.

Overall procedures of the system are illustrated in Section~\ref{sec:overall}. The voice recording starts when the user clicks the record button, and our system begins to print out the result.

\end{document}